%% file: main.tex
\documentclass[runningheads]{llncs}
\usepackage{amsmath} %added by XZ
\usepackage{centernot}
\usepackage{array}
\newcolumntype{P}[1]{>{\centering\arraybackslash}p{#1}}
\newcolumntype{M}[1]{>{\centering\arraybackslash}m{#1}}
\usepackage{amsfonts}%added by XZ
\usepackage{xcolor}%added by XZ
\usepackage{graphicx}

\usepackage[utf8]{inputenc}%added by XZ to have accent in the names like Barré ..

\usepackage{todonotes}
\usepackage{amsfonts, amsmath, amssymb, bbm}
\usepackage{comment} 
\usepackage{xcolor}
\usepackage{tikz}
\usepackage{hyperref}
\usetikzlibrary{arrows,shapes,snakes,automata, calc, chains, matrix, positioning, scopes, arrows.meta}

\usepackage{tabularx}
\usepackage{subcaption} 
\usepackage{url}
\usepackage{xpatch}
\usepackage{lscape}

\newcommand*\MyExp{{\mathbb E}}

\newcount\Comments  % 0 suppresses notes to selves in text
\usepackage{color}
\definecolor{darkgreen}{rgb}{0,0.5,0}
\definecolor{purple}{rgb}{1,0,1}
\newcommand{\kibitz}[2]{\ifnum\Comments=1\textcolor{#1}{#2}\fi}
\input{macro.tex}

\usepackage[absolute,overlay]{textpos}% by XZ: for the text in the top margin..

\begin{document}

%XZ: for the arXiv pre-print; to remove for the camera-ready version
\begin{textblock*}{20cm}(1cm,1cm)
\textcolor{red}{A preprint accepted by SafeComp2020. To appear in Springer LNCS series.}
\end{textblock*}
\title{A Safety Framework for Critical Systems Utilising Deep Neural Networks
}
\titlerunning{A Safety Framework for Deep Neural Networks}
\author{
Xingyu Zhao\inst{1} \and
Alec Banks\inst{2}\and
James Sharp\inst{2}\and
Valentin Robu\inst{1} \and
David Flynn \inst{1} \and
Michael Fisher \inst{3} \and
Xiaowei Huang\inst{3} 
}
\authorrunning{X. Zhao et al.}
\institute{
%School of Engineering and Physical Sciences,\\ 
Heriot-Watt University, Edinburgh, EH14 4AS, U.K.\\
\email{
%s.wang,
\{xingyu.zhao,v.robu,d.flynn\}@hw.ac.uk} \and
	Defence Science and Technology Laboratory, Salisbury, SP4 0JQ, U.K.\\
	\email{\{abanks,jsharp1\}@dstl.gov.uk}\\
\and
%Department of Computer Science, \\
University of Liverpool, Liverpool, L69 3BX, U.K.\\
\email{\{mfisher,xiaowei.huang\}@liverpool.ac.uk}
}
\maketitle              % typeset the header of the contribution
%
%\tableofcontents
\begin{abstract}
Increasingly sophisticated mathematical modelling processes from Machine Learning are being used to analyse complex data. However, the performance and explainability of these models within practical critical systems requires a rigorous and continuous verification of their safe utilisation. Working towards addressing this challenge, this paper presents a principled novel safety argument framework for critical systems that utilise deep neural networks. The approach allows various forms of predictions, e.g., future reliability of passing some demands, or confidence on a required reliability level. It is supported by a Bayesian analysis using operational data and the recent verification and validation techniques for deep learning. The prediction is conservative – it starts with partial prior knowledge obtained from lifecycle activities and then determines the worst-case prediction. Open challenges are also identified.
\end{abstract}
\section{Introduction}

Deep learning (DL) has been 
%shown successful in a number of tasks such as image classification, robotic control, and natural language processing etc. This motivates their application to 
applied broadly in industrial sectors including %both safety 
%critical sectors 
%and business critical applications, such as 
automotive, healthcare, aviation and finance.
To fully exploit the potential offered by DL, there is an urgent need to develop approaches to their certification in safety critical applications.  
%There is an urgent need to certify the safety of  critical systems, including their DL components.
%Safety cases have been used in a number of industry sectors to claim the safety of a software or hardware system, and seek the approval from the regulators. 
For traditional systems, 
%a typical safety case consists of a number of safety arguments together with various guidelines and standards \cite{bishop_methodology_2000,bloomfield_safety_2010}, see e.g., its application to medical device \cite{WeinstockTowardsan2009} and motor vehicle \cite{PH2010}. 
%These systems are usually deterministic and reactive, i.e., their behaviour is mostly predictable except for the hazards from the environment. In addition, these systems are usually developed from explicit functional requirement. 
safety analysis 
%for such conventional systems is 
%guided by well-established industry standards and supported by mature development processes and verification and validation (V\&V) techniques,
%and tools, 
has aided  engineers in \emph{arguing} that the system is sufficiently safe.
%see e.g., its application to medical device \cite{WeinstockTowardsan2009} and motor vehicle \cite{burton_making_2017}. 
%XZ: SafeComp audience should be quite familiar with the examples...
However, the deployment of DL in critical systems requires a thorough revisit of that analysis
%techniques to reflect new characteristics of machine learning (ML) techniques 
to reflect the novel characteristics of Machine Learning (ML) in general
\cite{BKCF2019,alves_considerations_2018,KKB2019}.

Compared with traditional systems, the behaviour of learning-enabled systems is much harder to predict, due to, \textit {inter alia}, their ``black-box'' nature and the lack of traceable functional requirements of their DL components. The ``black-box'' nature hinders the human operators in understanding the DL and makes it hard to predict the system behaviour when faced with new data. 
%This has led to the  research on interpretability \cite{Lipton2016} or explainable AI \cite{explainableAI}. 
The lack of explicit requirement traceability through to code implementation is only partially offset by learning from a dataset, which at best provides an incomplete description of the problem. 
%The data-driven nature leads to the existence of inherent hazards such as failures on a few desirable properties including robustness \cite{szegedy2014intriguing}, fairness \cite{barocas-hardt-narayanan}, and privacy \cite{ACGMMTZ2016} \xingyu{Will a ``failure'' of privacy has safety impact? As per our discussion, I am thinking how to position safety and other properties in a nice manner}.
These characteristics of DL increase apparent non-determinism \cite{johnson_increasing_2018}, which on the one hand emphasises the role of \textit{probabilistic measures} in capturing uncertainty, but on the other hand makes it notoriously hard to estimate the probabilities (and also the consequences) of critical failures. 
%To tackle these new challenges, the existing safety argument method needs to be adapted.

%Recently, progress has been made on formal verification \cite{HKWW2017} and 
%{katz2017reluplex,xiang2017output,GMDTCV2018,LM2017,wicker2018feature,RHK2018,wu2018game,ruan2018global}, 
%coverage-guided testing \cite{sun2018concolic}
%{PCYJ2017,sun2018testing,ma2018deepgauge,SHKSHA2019,sun2018concolicb}, etc., 
Recent progress has been made to support the Verification and Validation (V\&V) of DL, e.g., \cite{HKWW2017,sun2018concolic}.
Although these methods may provide evidence to support low-level claims, e.g., the local robustness of a deep neural network (DNN) on a given input, they are insufficient by themselves to justify overall system safety claims.
%due to their own limitations. 
%, but cannot be directly used to justify high-level claims such as the safety of the entire system or its DL components. 
%A comprehensive survey on existing methods for the safety and trustworthiness of deep learning can be found in \cite{Huangsurvey2018}. 
%
Here, we present a safety case framework for DL models which may in turn support higher-level system safety arguments.
%which run as components of a learning-enabled %system. 
We focus on DNNs
%, or more specifically convolutional neural networks (CNNs), 
that have been widely deployed
as, e.g., perception/control units of autonomous systems. Due to the page limit, we also confine the framework to DNNs that are fixed in the operation; this can be extended for online learning DNNs in future.
%in various applications, including perception and control units of autonomous systems.

We consider safety-related properties including reliability, robustness, %\cite{szegedy2014intriguing}, 
interpretability, fairness \cite{barocas-hardt-narayanan}, and privacy \cite{Abadi_2016}. In particular, we emphasise the assessment of DNN \textit{generalisation error} (in terms of inaccuracy), as a major reliability measure, throughout our safety case. We build arguments in two steps. The first is to provide initial confidence that the DNN's generalisation error is
%approaching its theoretical limit -- the Bayes error \cite{fukunaga_introduction_2013} -- 
%XZ: Actually we didn't show this in the later of the paper, so better omit this claim. and simply say it is bounded, I guess..
bounded, through the 
%desirable 
assurance activities conducted at each stage of its lifecycle, e.g., formal verification on the DNN robustness. The second step is to adopt \textit{proven-in-use/field-testing} arguments to boost the confidence and check whether the DNN is indeed sufficiently safe for the risk associated with its use in the system. 

The second step above is done in a statistically principled way via Conservative Bayesian Inference (CBI)  \cite{bishop_toward_2011,strigini_software_2013,zhao_assessing_2019}. 
%xiaowei: for space reason, too detailed and you did not discuss, so less informative. 
%, e.g., \cite{bishop_toward_2011,strigini_software_2013} given failure-free operations or \cite{zhao_assessing_2019} seeing few failures. 
CBI requires only \textit{limited and partial} prior knowledge of reliability, which differs from normal Bayesian analysis that usually assumes a \textit{complete} prior distribution on the failure rate.
%/probability. 
This has a unique advantage: partial prior knowledge is more convincing (i.e. constitutes a more realistic claim) and easier to obtain,
%in safety arguments, 
while complete prior distributions usually require extra assumptions and introduces optimistic bias. CBI allows many forms of prediction, e.g., posterior expected failure rate \cite{bishop_toward_2011}, future reliability of passing some demands \cite{strigini_software_2013} or a posterior confidence on a required reliability bound \cite{zhao_assessing_2019}. Importantly, CBI guarantees conservative outcomes: it finds the worst-case prior distribution yielding, say, a maximised posterior expected failure rate, and satisfying the partial knowledge.
We are aware that there are other extant dangerous pitfalls in safety arguments \cite{KKB2019,johnson_increasing_2018}, thus we also identify \textit{open challenges} in our proposed framework and map them onto on-going research.

%in the ML and software engineering communities.

The key contributions of this work are:

\textit{a)} A very first safety case framework for DNNs that mainly concerns \emph{quantitative} claims based on structured heterogeneous safety arguments.

\textit{b)} An initial idea of mapping DNN lifecycle activities to the reduction of decomposed DNN generalisation error that used as a primary reliability measure.

\textit{c)} Identification of open challenges in building safety arguments for quantitative claims, and mapping them onto on-going research of potential solutions.

Next, we present preliminaries.
%on safety cases, DNNs, and the notion of generalisation error in ML.
Sec.~\ref{sec_top_level_sc} provides top-level argument, and Sec.~\ref{sec_property_and_lifecycle} presents how CBI approach assures reliability. Other safety related properties are discussed in Sec.~\ref{sec-other-propreties}. We discuss related work in Sec.~\ref{sec-related} and conclude in Sec.~\ref{sec-conclusions}.

\section{Preliminaries}
\label{sec_preliminaries}

\subsection{Safety cases}
%Assurance cases are generally developed to support claims in areas such as safety, reliability, maintainability and security. These assurance cases are often called  by some more specific names, e.g. safety cases \cite{bishop_methodology_2000} and security cases \cite{knight_importance_2015}.
A safety case is a comprehensive, defensible, and valid justification of the safety of a system for a given application in a defined operating environment, thus it is a means to provide the grounds for confidence and to assist decision making in certification \cite{bloomfield_safety_2010}. %Safety cases have been widely used in the European safety community for decades to ensure system safety, e.g., they are mandatory in UK regulation for systems used in safety-critical industries like nuclear energy.
Early research in safety cases mainly focus on their formulation in terms of claims, arguments and evidence elements based on fundamental argumentation theories like the Toulmin model \cite{s_toulmin_uses_1958}. The two most popular notations are CAE \cite{bloomfield_safety_2010} and GSN \cite{kelly_arguing_1999}. In this paper, we choose the latter to present our safety case framework.

\begin{figure}[h!]
	\centering
	\includegraphics[width=1\textwidth]{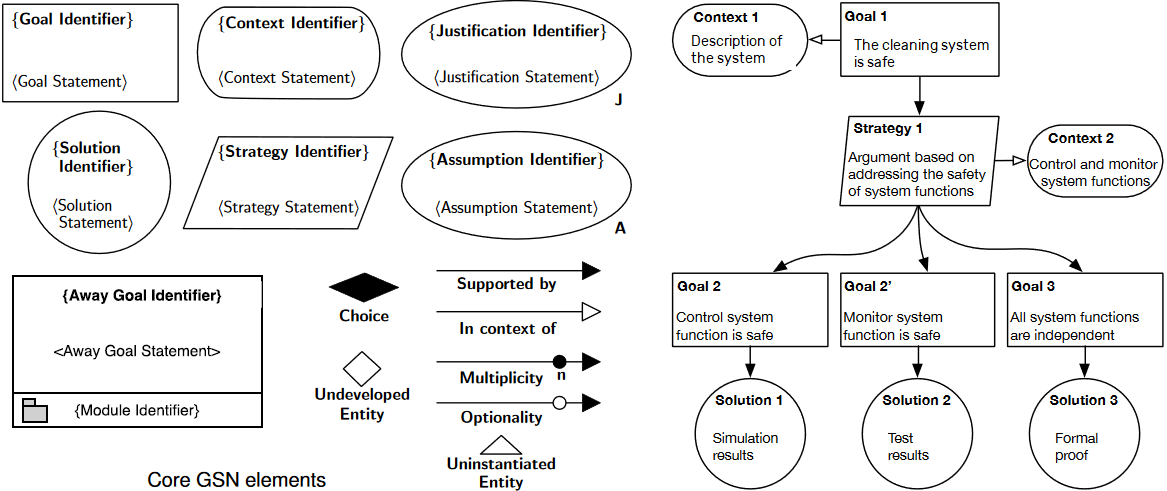}
	\caption{The GSN core elements and an example of using GSN.}
	\label{fig_GSN_example}
\end{figure}

%\vspace*{-\baselineskip}

Fig.~\ref{fig_GSN_example} shows the core GSN elements and a quick GSN example. Essentially, the GSN safety case starts with a top \textit{goal} (claim) which then is decomposed through an argument \textit {strategy} into sub-goals (sub-claims), and sub-goals can be further decomposed until being supported by \textit{solutions} (evidence). %The decomposition strategy can be either inductive (i.e., incomplete) or deductive (i.e., complete) \cite{alves_considerations_2018}. For an inductive strategy, additional analysis is required to ensure that residual risks are mitigated. 
A claim may be subject to some \emph{context} or \emph{assumption}. An \emph{away goal} repeats a claim presented
in another argument module.
A description on all GSN elements used here can be found in \cite{kelly_arguing_1999}.

\subsection{Deep neural networks and lifecycle models}

%, $O$ an operational profile in which $\network$ runs,  and $\phi$ a formula representing a required safety specification.
%Let $\networks$ be the (possibly infinite) set of networks of a given architecture and $\contexts$ be the (possibly infinite) set of contexts.
%
%The size of $|\networks|$ is also called the capacity of the architecture. 
%For a learning model, there are methods to evaluate its  capacity, for example the VC dimensions \cite{Vapnik2015}. 
%
%\paragraph{Datasets} 
%
Let $(X,Y)$ be the training data, where $X$ is a vector of inputs and $Y$ is a 
%corresponding 
vector of outputs such that $|X|=|Y|$. 
%Similarly, we also have testing dataset $(X',Y')$ with $|X'|=|Y'|$. 
%We may use $x$ and $y$ to range over $X\cup X'$ and $Y\cup Y'$, respectively. 
%
Let $\inputdomain$ be the input domain and $\outputdomain$ be the set of labels. Hence, %$X\cup X'\subset \inputdomain$ and $Y\cup Y'\subset \outputdomain$.
$X\subset \inputdomain$.
We may use $x$ and $y$ to range over $\inputdomain$ and $\outputdomain$, respectively. Let $\network$ be a DNN of a given architecture.
A network $\network:\inputdomain\rightarrow \dist(\outputdomain)$ can be seen as a function mapping from $\inputdomain$ to probabilistic distributions over $\outputdomain$. That is, $\network(x)$ is a probabilistic distribution, which assigns for each possible label $y\in \outputdomain$ a probability value $(\network(x))_y$. We let $f_\network:\inputdomain\rightarrow \outputdomain$ be a function such that for any $x\in \inputdomain$, 
$f_\network(x) = \arg\max_{y\in \outputdomain}\{(\network(x))_y\}$, i.e. $f_\network(x)$ returns the classification label.
The network is trained with a parameterised learning algorithm, in which there are (implicit) parameters representing e.g., the number of epochs, the loss function, the learning rate, the optimisation algorithm, etc. 

%\paragraph{Learning Algorithm} We use  $\trainingAlgorithm_{(X,Y)}$ to represent a parameterised learning algorithm, in which there are (implicit) parameters representing e.g., the number of epochs, the loss function, the learning rate, the optimisation algorithm, etc. 
%Let $\networks$ be the (possibly infinite) set of networks of a given architecture. Then, $\trainingAlgorithm_{(X,Y)}:\networks \rightarrow \networks$, mapping from a network to another network, represents a learning procedure. 

%\paragraph{Lifecycle Model} 

A comprehensive ML \textit{Lifecycle Model} can be found in \cite{ashmore_assuring_2019}, which identifies assurance desiderata for each stage, and reviews existing methods that contribute to achieving these desiderata. In this paper, we refer to a simpler lifecycle model that includes several phases: initiation, data collection, model construction, model training, analysis of the trained model, and run-time enforcement. 
%Moreover, one may consider safety architecture where multiple components apply together to improve the safety performance, as the \emph{defence in depth design} to be discussed in Section~\ref{sec_top_level_sc}.   

\subsection{Generalisation error}

Generalisability requires that a neural network  works well on all possible inputs in $\inputdomain$, although it is only trained on the training dataset $(X,Y)$. 

%A direct consequence of a good generalisability is that the network will not be over-fitting, which says that  
%a network may fit well to a particular set of data samples but fail to perform as well on additional data. 

\begin{definition} %[Generalisability]
	Assume that there is a ground truth function $f: \inputdomain\rightarrow \outputdomain$ and a probability function $O_p: \inputdomain\rightarrow [0,1]$ representing the operational profile. A network $\network$ trained on $(X,Y)$ has a generalisation error: 
	\begin{equation}
G^{0-1}_\network = \sum_{x\in \inputdomain}  {\bf 1}_{\{f_\network(x) \neq f(x)\}}  \times O_p(x)
\label{eq_gen_error_01}
\end{equation}
where ${\bf 1}_{\tt S}$ is an indicator function -- it is equal to 1 when {\tt S} is true and 0 otherwise.
%	Moreover, $\network$ has a better generalisability than  $\network'$ if $G^{0-1}_\network < G^{0-1}_{\network'}$
\end{definition}

We use the notation $O_p(x)$ to represent the probability of an
%future 
input $x$ being selected, which aligns with the \textit{operational profile} notion \cite{musa_operational_1993} in software engineering. 
Moreover, we use 0-1 loss function (i.e., assigns value 0 to loss for a correct classification and 1 for an incorrect classification) so that, for a given $O_p$, $G^{0-1}_\network$ is equivalent to the reliability measure \textit{pfd} (the expected probability of the system failing on a random demand) defined in the safety standard IEC-61508. %\cite{iec_61508_2010}.
A ``frequentist'' interpretation of \textit{pfd} is that it is the limiting relative frequency of demands for which the DNN fails in an infinite sequence of independently selected demands \cite{zhao_modeling_2017}. The primary safety measure we study here is \textit{pfd},
%, a 
%jargon of notation studied in 
%the safety and reliability community. 
%Because it 
which is equivalent to the generalisation error $G^{0-1}_\network$ in \eqref{eq_gen_error_01}. Thus, we may use the two terms interchangeably in our safety case, depending on the context.

\section{The Top-level Argument}
\label{sec_top_level_sc}

\begin{figure*}[htb]
	\centering
	\includegraphics[width=0.85\textwidth]{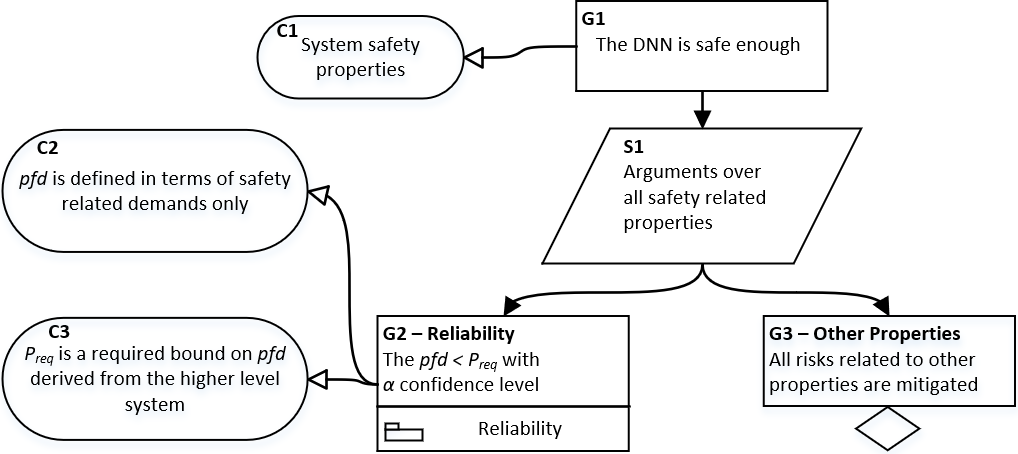}
	\caption{The top-level safety argument}
	\label{fig_top_level}
\end{figure*}

Fig.~\ref{fig_top_level} gives a top-level safety argument for the top claim \textbf{G1} -- the DNN is sufficiently safe.
We first argue \textbf{S1}: that all safety related properties are satisfied. The list of all properties of interest for the given application can be obtained by utilising the Property Based Requirements (PBR) \cite{Micouin2008} approach. 
The PBR method is a way to specify requirements as a set of properties of system objects in either structured language or formal notations. PBR is recommended in \cite{alves_considerations_2018} as a method for the safety argument of autonomous systems. Without the loss of generality, in this paper, we focus on the major quantitative property: reliability (\textbf{G2}). Due to space constraints, other properties: interpretability, robustness, etc. are discussed in Sec.~\ref{sec-other-propreties} but remain an undeveloped goal (\textbf{G3}) here.

%and iterated over all properties). 
More properties that have a safety impact can be incorporated in the framework as new requirements emerge from, e.g., ethical aspects of the DNN.
%Then we support this goal by splitting the product-based arguments and process-based arguments as suggested in \cite{the_assurance_case_working_group_goal_2018} and also proposed in \cite{schwalbe_concept_2020}.

%\vspace*{-\baselineskip}

Despite the controversy over the use of probabilistic measures (e.g., \textit{pfd}) for the safety of conventional software systems \cite{littlewood_validation_2011}, we believe probabilistic measures %become 
%even more important crucial
are useful 
when dealing with ML systems since arguments involving their inherent uncertainty are naturally stated in probabilistic terms.

%e.g., the \textbf{C2} for \textit{pfd} in fig.~\ref{fig_top_level}.

%When considering \textbf{G2}, how exactly
%it is appropriate 
%to set a reliability goal 
%(\textbf{G2}) 

Setting a reliability goal (\textbf{G2})
for a DNN varies from one application to another. 
Questions we need to ask include: (i) What is the appropriate reliability measure? (ii) What is the quantitative requirement stated in that reliability measure? (iii) How can confidence 
be gained in that reliability claim?

Reliability of safety critical systems, as a probabilistic claim, will be about the probabilities/rates of occurrence of failures that have safety impacts, e.g., a dangerous misclassification in 
a DNN. Generally, systems can be classified as either: continuous-time systems that are being continuously operated in the active control of some process; or on-demand systems, which are only called upon to act on receipt of discrete demands. Normally we study the failure rate (number of failures in one time unit) of the former (e.g., flight control software) and the probability of failure per demand (\textit{pfd}) of the latter (e.g., the emergency shutdown system of a nuclear plant).
%Other reliability measure may also make sense, e.g., (i) the probability of surviving a mission for systems launching a spacecraft or (ii) the probability of seeing fatalities per mile as studied in \cite{kalra_driving_2016,zhao_assessing_2019} for self-driving cars. 
In this paper, we focus on \textit{pfd} which aligns with DNN classifiers for perception, where demands are e.g., images from cameras.

Given the fact that most safety critical systems adopt a \textit{defence in depth design} with safety backup channels \cite{littlewood_reasoning_2012}, the required reliability ($p_{\mathit{req}}$ in \textbf{G2}) should be derived from the higher level system, e.g., a 1-out-of-2 (1oo2) system in which the other channel could be either hardware-only, conventional software-based, or another ML software. The required reliability of the whole 1oo2 system may be obtained from regulators or compared to human level performance (e.g., a target of 100 times safer than average human drivers, as studied in \cite{zhao_assessing_2019}).
%liu_how_2019
We remark that deriving a required reliability for individual channels to meet the whole 1oo2 reliability requirement is still an open challenge due to the dependencies among channels \cite{littlewood_conceptual_1989,littlewood_conservative_2013} (e.g., a ``hard'' demand is likely to cause both channels to fail). That said, there is ongoing research towards rigorous methods to decompose the reliability of 1oo2 systems into those of individual channels which may apply and provide insights for future work, e.g., \cite{bishop_conservative_2014} for 1oo2 systems with one hardware-only and one software-based channels, \cite{littlewood_reasoning_2012,zhao_modeling_2017} for a 1oo2 system with one possibly-perfect channel, and  \cite{chen_diversity_2016} utilising fault-injection technique. In particular, for systems with duplicated DL channels, we note that there are similar techniques, e.g., (i) ensemble method \cite{Ponti2011}, where a set of DL models run in parallel and the result is obtained by applying a voting protocol; (ii) simplex architecture \cite{Sha2001}, where there is a main classifier and a safer classifier, with the latter being simple enough so that its safety can be formally verified. Whenever confidence of the main classifier is low, the decision making is taken over by the safer classifier; the safer classifier can be implemented with e.g., a smaller DNN.

As discussed in \cite{bishop_toward_2011}, the reliability measure, \textit{pfd}, 
%discussed above 
concerns system behaviour subject to \textit{aleatory} uncertainty (``uncertainty in the world'').
%that 
%called \textit{aleatory} uncertainty.
%in the jargon. 
On the other hand,  \textit{epistemic} uncertainty concerns the uncertainty in 
%While, there also exists 
the ``beliefs about the world''. In our context, it is about the human assessor's \textit{epistemic} uncertainty of the reliability claim obtained through  assurance activities. For example, we may not be \textit{certain} whether a claim  -- the \textit{pfd} is smaller than $10^{-4}$ -- is true due to our imperfect 
%human knowledge 
understanding
about the assurance activities. All assurance activities in the lifecycle with supportive evidence would increase our \textit{confidence} in the reliability claim, whose formal quantitative treatment has been proposed in \cite{bloomfield_confidence:_2007,littlewood_use_2007}.
%littlewood_use_2007
Similarly to the idea proposed in \cite{strigini_software_2013}, we argue that all ``process'' evidence generated from the DNN lifecycle activities provides initial confidence of a desired \textit{pfd} bound.
%(generalisation error in \eqref{eq_gen_error_01}). 
Then the confidence in a \textit{pfd} claim is acquired incrementally through operational data of the trained DNN via CBI -- which we describe next.

\section{Reliability with Lifecycle Assurance}
\label{sec_property_and_lifecycle}

\subsection{CBI utilising operational data}

In Bayesian reliability analysis, assessors normally have a prior distribution of \textit{pfd} (capturing the \textit {epistemic} uncertainties), and update their beliefs -- the prior distribution -- by using evidence of the observed operational data. Given the safety-critical nature, the systems under study will typically see \textit{failure-free} operation or very \textit{rare failures}. Bayesian inference based on such non or rare failures may introduce dangerously optimistic bias if using a \textit{Uniform} or \textit{Jeffreys prior}
%(as suggested by regulatory guidance like \cite{atwood2003handbook})
which describes not only one's prior knowledge, but adds extra, unjustified assumptions \cite{zhao_assessing_2019}. Alternatively, CBI is a technique, first described in \cite{bishop_toward_2011}, 
%that 
which
applied Bayesian analysis with only \textit{partial} prior knowledge;
%which is much easier to be obtained/justified in practice than a \textit{complete} prior distribution and can avoid optimistic bias.
by partial prior knowledge, we mean the following typical forms: 
	\begin{itemize}
		\item $\MyExp[\textit{pfd}] \leq m$: the prior mean \emph{pfd} cannot be worse than a stated value;
		\item $Pr(\textit{pfd}\leq\epsilon)=\theta$: a prior confidence bound on \emph{pfd};
		\item $Pr(\textit{pfd}=0)=\theta$: a prior confidence in the perfection of the system;
		\item $\MyExp[(1-\textit{pfd})^n] \geq  \gamma$: prior confidence in the reliability of passing $n$ tests.
	\end{itemize} 
	%which 

These can be used by CBI either solely or in combination (e.g., several confidence bounds). The partial prior knowledge is far from a complete prior distribution, thus it is easier to obtain from DNN lifecycle activities (\textbf{C4}). For instance, there are studies on the generalisation error bounds, based on how the DNN was constructed, trained and verified \cite{he_control_2019,bagnall_certifying_2019}.
%\cite{jakubovitz_generalization_2019,he_control_2019,bagnall_certifying_2019}. 
We present examples on how to obtain such partial prior knowledge (\textbf{G6})
%in Fig.~\ref{fig_CBI_case} and \ref{fig_partial_prior_knowledge}
 using evidence, e.g. from formal verification on DNN robustness, in the next section. CBI has also been investigated for various objective functions	with a ``posterior'' flavour:
\begin{itemize}
	\item $\MyExp[\textit{pfd}\mid\mbox{pass }n\mbox{ tests}] $: the posterior expected \emph{pfd} \cite{bishop_toward_2011}; 
	\item $Pr(\textit{pfd} \leq p_{req}\mid k \mbox{ failures }\mbox{in }n\mbox{ tests})$: the posterior confidence bound on \emph{pfd} \cite{zhao_modeling_2017,zhao_assessing_2019}; the $p_{req}$ is normally a small \emph{pfd}, stipulated at higher level;
%	\item  $Pr(\textit{pfd}=0\mid\mbox{pass }n\mbox{ tests})$: the posterior probability of perfection in \cite{zhao_conservative_2015};
	\item $\MyExp[(1-\textit{pfd})^t\mid \mbox{pass }n\mbox{ tests}] $: the future reliability of passing $t$ demands in \cite{strigini_software_2013}.
\end{itemize}

\begin{example}
\label{exp_CBI_example}
In Fig.~\ref{fig_CBI_example}, we plot a set of numerical examples based on the CBI model in  \cite{strigini_software_2013}. It describes the following scenario: the assessor has $\theta$ confidence that the software \textit{pfd} cannot be worse than $\epsilon$ (e.g., $10^{-4}$ according to SIL-4), then after $n$ failure-free runs (the x-axis), the future reliability of passing $t$ demands is shown on the y-axis. We may observe that stronger prior beliefs (smaller $\epsilon$ with larger $\theta$) and/or larger $n/t$ ratio allows higher future reliability claims.  
\begin{figure}[h!]
	\centering
	\includegraphics[width=1\textwidth]{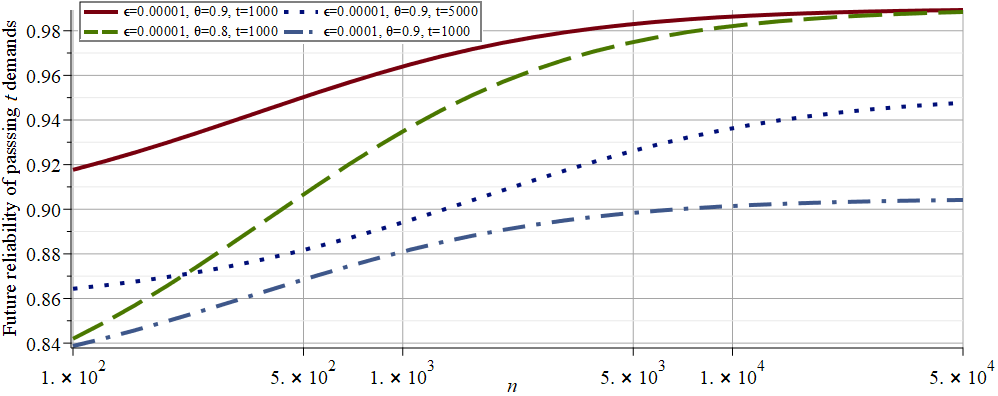}
	\caption{Numerical examples based on the CBI model in \cite{strigini_software_2013}.}
	\label{fig_CBI_example}
\end{figure}
\end{example}

Depending on the objective function of interest (\textbf{G2} is an example of a posterior confidence bound) and the set of partial prior knowledge obtained (\textbf{G6}), we choose a corresponding CBI model\footnote{CBI is an ongoing research proving theorems for some combinations of objective functions and partial prior knowledge. There are combinations haven't been investigated, which remains as open challenges.} for \textbf{S2}. 
%That said, 
Note,
we also need to explicitly assess the impact of CBI model assumptions (\textbf{G5}). Published CBI theorems abstract the stochastic failure process as a sequence of independent and identically distributed (i.i.d.) Bernoulli trials given the unknown \textit{pfd}, and assume the operational profile is constant \cite{bishop_toward_2011,strigini_software_2013,zhao_assessing_2019}. 
Although we identify how to justify/relax those assumptions as open challenges, we note some promising ongoing research:

\textit{a)} The i.i.d. assumption means a constant \textit{pfd} (a frozen system in an unchanging environment), which may not hold
%when seeing 
for
a system update or deployment in a new environment. In \cite{littlewood_reliability_2020}, the CBI is extended to a \textit{multivariate} prior distribution case, which deals with scenarios of a changing \textit{pfd}. The multivariate CBI may provide the basis of arguments for online learning DNNs.

\textit{b)} The effect of assuming independence between successive demands has been studied, e.g., \cite{strigini_testing_1996,galves_rare_1998}.
It is believed that the effect is negligible given non or rare failures; note this requires further
%, which, however, needs more 
(preferably conservative) studies.

\textit{c)} The changes to the
%ing of 
operational profile is a major challenge for all proven-in-use/field-testing safety arguments \cite{KKB2019}. Recent research \cite{bishop_deriving_2017} provides a novel conservative treatment for the problem, which can be retrofitted for CBI.
	%(an unpublished work to appear).

\begin{figure}[htb]
	\centering
	\includegraphics[width=0.85\textwidth]{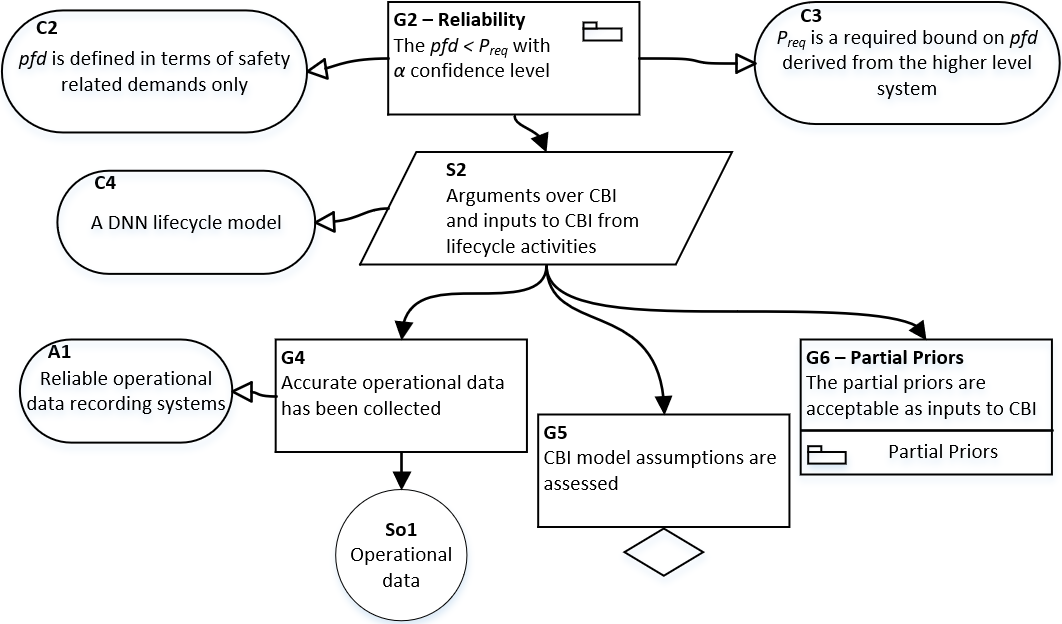}
	\caption{The CBI safety argument.}
	\label{fig_CBI_case}
\end{figure}

The safety argument via CBI is presented in Fig.~\ref{fig_CBI_case}. In summary, we collect a set of partial prior knowledge
%(e.g., initial confidence bounds on \textit{pfd}) 
from various lifecycle activities, then boost our posterior confidence in a reliability claim of interest 
%via
through operational data, in a conservative Bayesian 
%way. 
manner.
We believe this aligns with the practice of applying management systems in reality -- a system is built 
%try the best to 
%build,
%
%what they consider to be,
%
%a %very 
%reliable system with
%initial confidence that it is safe to deploy
with claims of sufficient confidence that it may be deployed;
%, then 
these claims are then independently assessed 
%boost their confidence that it is indeed safe enough via good operation records.
to confirm said confidence is justified. %in certifying that it is suitably safe to deploy.
Once deployed, the system safety performance is then monitored for continuing validation of the claims. Where there is insufficient evidence systems can be fielded with the risk held by the operator, but that risk must be minimised through operational restrictions. As confidence then grows these restrictions may be relaxed.  

%\vspace*{-\baselineskip}

\subsection{Partial prior knowledge on the generalisation error}

Our novel CBI safety argument for the reliability of DNNs is essentially inspired by the idea proposed in \cite{strigini_software_2013} for conventional software, in which the authors seek prior confidence in the (quasi-)perfection of the software from ``process'' evidence like formal proofs, and effective development activities. In our case, to make clear the connection between lifecycle activities and their contributions to the generalisation error, we 
%firstly 
decompose the generalisation error into three: 
%\xingyu{Is this accurate? $\network$ is ambiguous.. One is the learnt DNN, the other is a variable representing some DNN... But i see why you change the orignal one. Now I tend to change the eq. 1 as a function of the learnt DNN... like \url{http://www.shivani-agarwal.net/Teaching/CIS-520/Spring-2018/Lectures/Reading/error-bounds-decompositions.pdf}}
%for some insights: We use the 0-1 loss with a fixed the operational profile (indicated by the upper index $^{0-1}$ and lower index $_{Op}$ respectively in \eqref{eq_decomp_ge}) as an example, then treat the generalisation error as a function of a DNN classifier $\network_s$ learnt from the set of networks $\networks$ for a given architecture:
\begin{equation}
\label{eq_decomp_ge}
G^{0-1}_\network =
\underbrace{G^{0-1}_{\network} -\inf_{\network \in \networks}G^{0-1}_\network}_\text{Estimation error of $\network$}
+
\underbrace{\inf_{\network \in \networks}G^{0-1}_\network-G^{0-1,*}_{f,(X,Y)}}_\text{Approximation error of $\networks$}
+\underbrace{G^{0-1,*}_{f,(X,Y)}}_\text{Bayes error}
\end{equation}

%\xingyu{XH, please check if I am saying something wrong below...}

\textit{a)} The \textit{Bayes error} is the lowest and irreducible error rate over all possible classifiers for the given classification problem \cite{fukunaga_introduction_2013}. It is non-zero if the true labels are not deterministic (e.g., an image being labelled as $y_1$ by one person but as $y_2$ by others), thus intuitively it captures the uncertainties in the dataset $(X,Y)$ and true distribution $f$ when aiming to solve a real-world problem with DL. We estimate 
this error
(implicitly) 
%it 
at the \textbf{initiation} and \textbf{data collection} stages in activities like: necessity consideration and dataset preparation etc.

\textit{b)} The \textit{Approximation error of $\networks$} measures how far the best classifier in $\networks$ is from the overall optimal classifier, after isolating the Bayes error. The set $\networks$ is determined by the architecture of DNNs (e.g., numbers of layers
% and activation functions
), thus lifecycle activities at the \textbf{model construction} stage are used to minimise this error.
%, essentially.

\textit{c)} The \textit{Estimation error of $\network$} measures how far the learned classifier $\network$
is from the best classifier in $\networks$. Lifecycle activities at the \textbf{model training} stage 
%after the DNN being constructed 
essentially aim to reduce this error, i.e., 
performing
%doing
optimisations 
%in 
of 
the set $\networks$.

Both the Approximation and Estimation errors are reducible. We believe, the \emph{ultimate goal} of all lifecycle activities is to reduce the two errors to 0, especially for safety-critical DNNs. This is analogous to the ``possible perfection'' notion of traditional software as pointed to by Rushby and Littlewood \cite{littlewood_reasoning_2012,rushby_software_2009}. That is, assurance activities, e.g., performed in support of DO-178C, can be best understood as developing evidence of possible perfection -- a confidence in $\mathit{pfd}=0$. Similarly, for safety critical DNNs, we believe ML lifecycle activities should be considered as aiming
to train a ``possible perfect'' DNN in terms of the reducible Approximation and Estimation errors. Thus, we may have some confidence that the two errors are both 0 (equivalently, a prior confidence in the irreducible Bayes error since the other two are 0, that can be used by CBI), which indeed is supported by on-going research into 
finding globally optimised DNNs \cite{du_gradient_2018}. Meanwhile, on the \textbf{trained model}, V\&V also provides prior knowledge as shown in Example~\ref{example_robustness} 
below, and \textbf{online monitoring} continuously validates the assumptions for the prior knowledge being obtained.

%are to compute the combined estimation and approximation error, 
%and the \textbf{run-time enforcement} is to reduce the overall generalisation error. XZ: since we confine to fixed DNN, online learning is not considered here. ..
%As discussed, every stage in the ML lifecycle is to reduce, or evaluate, one or more of these three errors. 
%For statistical risk assessment, we will collect partial prior knowledge from these stages. 

\begin{figure}[tbh]
	\centering
	\includegraphics[width=0.8\textwidth]{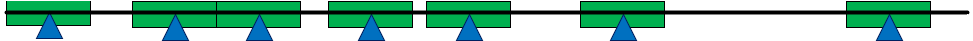}
	\caption{Formal verification on DNN robustness in an one-dimensional space.}
	\label{fig_illustrate_robust_veri}
\end{figure}

\begin{example}
\label{example_robustness}
We present an illustrative example on how to obtain a prior confidence bound on the generalisation error from formal verification of DNN robustness \cite{ruan2018global,HKWW2017}. \textit{Robustness} requires that the decision making of a neural network cannot be drastically changed due to a small perturbation on the input. Formally, given a real number $d > 0$ and a distance measure $\distance{\cdot}{p}$, for any input $x\in \inputdomain$, we have that, $f_\network(x) = f_\network(x')$ whenever $\distance{x'-x}{p}\leq d$.

Fig.~\ref{fig_illustrate_robust_veri} shows an example of the robustness verification in a one-dimensional space. Each blue triangle represents an input $x$, and the green region around each input $x$ represents all the neighbours, $x'$ of $x$,
which satisfy $\distance{x'-x}{p}\leq d$ and $f_\network(x) = f_\network(x')$. Now if we assume $Op(x)$ is uniformly distributed (an assumption for illustrative purposes which can be relaxed for other given $Op(x)$ distributions), the generalisation error has a lower bound -- the chance that the next randomly selected input does not fall into the green regions. That is, if 
$\epsilon$ denotes the ratio of the length not being covered by the green regions to the total length of the black line, then $G^{0-1}_\network \leq \epsilon$. 
This
said, we cannot be certain about the bound $G^{0-1}_\network \leq \epsilon$ due to assumptions like: (i) The formal verification tool itself is perfect, which may not hold; %\cite{beyer_software_2017}; 
(ii) Any neighbour $x'$ of $x$ %(defined by $\distance{x'-x}{p}\leq d$) 
has the same ground truth label of $x$. For a more comprehensive list, cf. \cite{burton_confidence_2019}. Assessors need to capture the doubt (say $1-\theta$) in those assumptions, which leads to:
%a prior confidence bound of 
\begin{equation}\label{equation-robustness}
Pr(G^{0-1}_\network \leq \epsilon)=\theta . 
\end{equation}
\end{example}

\begin{figure}[tbh]
	\centering
	\includegraphics[width=0.85\textwidth]{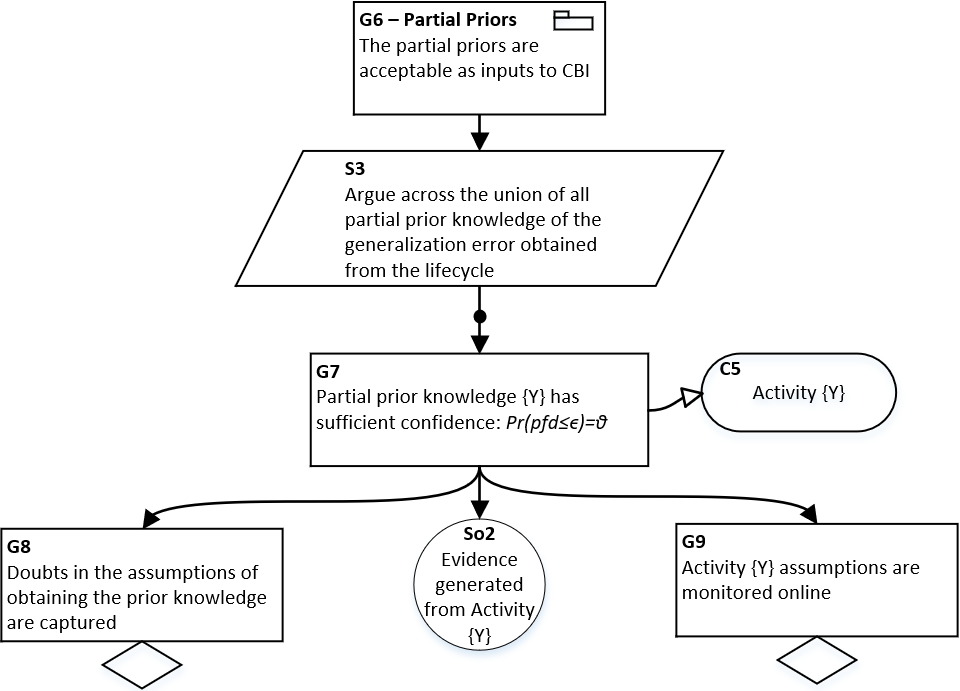}
	\caption{A template of safety arguments for obtaining partial prior knowledge.}
	\label{fig_partial_prior_knowledge}
\end{figure}

So far, we have presented an instance of the safety argument template in Fig.~\ref{fig_partial_prior_knowledge}. The solution \textbf{So2} is the formal verification result showing $G^{0-1}_\network \leq \epsilon$, and \textbf{G8} in Fig.~\ref{fig_partial_prior_knowledge} quantifies the confidence $\theta$ in that result. It is indeed an open challenge to rigorously develop \textbf{G8} further, which may involve scientific ways of eliciting expert judgement \cite{ohagan_uncertain_2006} and systematically collecting process data (e.g., statistics on the reliability of verification tools). However, we believe this challenge -- evaluating confidence in claims, either quantitatively or qualitatively (e.g., ranking with low, medium, high), explicitly or implicitly -- is a fundamental problem for all safety case based decision-makings
%, see relevant discussions in e.g., 
\cite{denney_towards_2011,bloomfield_confidence:_2007,zhao_new_2012,wang_confidence_2017}, rather than a specific problem of our framework.
%wang_confidence_2017
%zhao_new_2012

The sub-goal \textbf{G9} represents the mechanism of online monitoring on the validity of offline actives, e.g., validating the environmental assumptions used by offline formal verifications against the real environment at runtime
%in 
\cite{ferrando_verifying_2018}.

%\vspace*{-\baselineskip}
%\vspace*{-\baselineskip}

%\vspace*{-\baselineskip}
%\vspace*{-\baselineskip}

\section{Other Safety Related Properties}\label{sec-other-propreties}

So far we have seen a reliability-centric safety case for DNNs. Recall that, in this paper, reliability is the probability of misclassification (i.e. the generalisation error in \eqref{eq_gen_error_01}) that has safety impacts. % e.g., a misclassification of a traffic sign of ``turn right'' into ``turn left''. 
However, there are other DNN safety related properties concerning 
risks not directly caused by a misclassification, like interpretability, fairness, and privacy; discussed as follows. %, which we will discuss as follows.

\textit{Interpretability} is about an explanation procedure to present an interpretation of a single decision within
the overall model in a way
that is easy for humans to understand. 
%EU General Data Protection Regulation (GDPR) regulated that any significant or legally related decision needs to be explainable. 
There are different explanation techniques aiming to work with different objects, see \cite{Huangsurvey2018} for a survey. Here we take the instance explanation as an example -- the goal is to find another representation $\explain(f_\network,x)$ of an input $x$, with the expectation that $\explain(f_\network,x)$ carries simple, yet essential, information that can help the user understand the decision $f_\network(x)$. We use $f(x)\Leftrightarrow\explain(f_\network,x)$ to denote that the explanation is consistent with a human's explanation in $f(x)$. Thus, similarly to \eqref{eq_gen_error_01}, we can define a probabilistic measure for the instance-wise interpretability: 
\begin{equation}
I_\network = \sum_{x\in \inputdomain}  (f(x) \centernot\iff \explain(f_\network,x))  \times O_p(x)
\label{eq_interpretability}
\end{equation}

Then similarly as the argument for reliability, we can do statistical inference with the probabilistic measure $I_\network$. For instance, as in
Ex.~\ref{example_robustness}, we (i) firstly define the robustness of explanations in norm balls, measuring the percentage of space that has been verified as a bound on $I_\network$, (ii) then estimate the confidence of the robust explanation assumption and obtain a prior confidence in interpretability, (iii) finally Bayesian inference is applied with runtime data.

%Robustness, discussed in Example~\ref{example_robustness}, may not be the only property required when evaluating the reliability of a learning-enabled system.  In the following, we discuss several other critical properties of DL, and suggest how to obtain their respective partial prior knowledge. 

\begin{comment}
Alternatively, we may sample a number of inputs $x$, compute the empirical failure rate $\textit{pfd}_{empirical}$ in achieving consistent explanation, and use Chebyshev's inequality to establish expression like 	
\begin{equation}
Pr(
|G^{0-1}_\network - \textit{pfd}_{empirical}| \leq \epsilon)\geq \gamma
\end{equation}
for any number $\epsilon > 0$, where $\gamma$ is an expression parameterised over the number of verified inputs and the parameters of function $f_\network$.  

\end{comment}

\textit{Fairness} requires that, when using DL to predict an output, the prediction remains unbiased with respect to some protected features. For example, a financial service company may use DL to decide whether or not to provide loans to an applicant, and it is expected that such decision should not rely on sensitive features such as race and gender. \textit{Privacy} is used to prevent an observer from determining whether or not a sample was in the model's training dataset, when it is not allowed to observe the dataset directly. Training methods such as \cite{Abadi_2016} %at the model training phase 
have been applied to pursue %$\epsilon$-
differential privacy.

\begin{comment}

There are several variants of fairness \cite{Gajane2017,CHKV2018}, including e.g., fairness through unawareness, counterfactual fairness, and equality of opportunities. Here we consider counterfactual fairness.  

\begin{example}
Let $\varY$ be the decision and $\varZ$ be the sensitive features, we ask for the independence between them in the joint distribution  $Pr(\varX,\varY,\varZ)$, where $\varX$ is a random variable over the input domain. We can quantify the independence between $\varY$ and $\varZ$ with their correlation coefficient $\rho_{\varY,\varZ} \in [-1,1]$. Then, given a sample set of inputs $X'$, we can define failure on $X'$ as $\rho_{\varY,\varZ}(X') > \epsilon$, for some pre-specified threshold $\epsilon$ that represents an acceptable level of dependence. 

Equation (\ref{eq_gen_error_01}) can be generalised, and for partial prior knowledge, we can have 
\begin{equation}\label{expression-fairness}
Pr(
\textit{pfd}=0)= \theta
\end{equation}
after verifying a number of sets of inputs without experiencing any failure. Here, we ask for $\textit{pfd}=0$ to denote that any failure is deemed unacceptable. This is because we already set the threshold $\epsilon$ for the correlation coefficient. 
\end{example}

The open challenge will be on the computation of $\theta$, which is the percentage of verified sets $X'$ among the powerset of inputs $2^{\inputdomain}$. Even for discrete input space, it will be computationally hard to have a high percentage of input sets verified. 

\end{comment}

The lack of fairness or privacy may cause not only a significant monetary loss but also ethical issues. 
%The latter 
Ethics
has been regarded as a long-term challenge
%of 
for AI safety.
%, as opposed to the reliability which is a short-term challenge. 
%For sure, we are not enumerating all safety related properties as more to emerge as the research on DNNs continues. 
For these properties, we believe the general methodology suggested here still works -- we first introduce bespoke probabilistic measures according to their definitions, obtain prior knowledge on the measures from lifecycle activities, then conduct statistical inference 
during the continuous monitoring of the operation. 
%with continuously monitoring during the operation.

\begin{comment}

We consider $\epsilon$-differential privacy which says that the privacy of an element (e.g., a person) cannot be compromised by a statistical release if their data are not in the dataset.

Equation (\ref{eq_gen_error_01}) can be generalised, and for the prior knowledge, we can follow similar approach as in fairness to have an expression like Equation (\ref{expression-fairness}).

The open challenge is similar to that of fairness, except that it can be harder due to the fact that every verification on $X_1$ and $X_2$ may require a training of two neural networks. There are attack methods such as membership inference attack and attribute attack, but it needs to establish how to utilise such attacks for the computation of probabilistic values in Equation (\ref{definition-privacy}). 
\end{comment}

\section{Related Work}
\label{sec-related}

%Here we firstly review research moving towards safety cases of the more general AI/autonomous systems which include learning components. Then we discuss some domain-specific safety cases framework for ML.

%XZ: cite{asaadi_towards_2019} this work needs to be added here later with, since it is our rare ally that actually loves ``quantitative'' assurance....

%The paper 
Alves \textit{et.al} \cite{alves_considerations_2018} present a comprehensive discussion on the aspects that need to be considered when developing a safety case for increasingly autonomous systems that contain ML components. In \cite{BKCF2019}, a safety case framework with specific challenges for ML is proposed.
%The work 
\cite{SS2020} reviews available certification techniques from the aspects of lifecycle phases, maturity and applicability to different types of ML systems. In \cite{KKB2019}, safety arguments that are being widely used for conventional systems -- including conformance to standards, proven in use, field testing, simulation and formal proofs --  are recapped for  autonomous systems with discussions on the potential pitfalls. Similar to our CBI arguments that exploit operational data, \cite{matsuno_tackling_2019,ishikawa_continuous_2018} propose utilising continuously updated arguments to monitor the weak points and the effectiveness of their countermeasures. The work \cite{asaadi_towards_2019} identifies applicable quantitative measures of assurance for learning-enabled components.
% The work \cite{asaadi_towards_2019} is also interested in quantitative claims in safety assurance arguments, after identifying the applicable quantitative measures of assurance, and characterising the associated uncertainty probabilistically.

%Specifically for
Regarding 
the safety of automated driving, \cite{SS2020b,rudolph_consistent_2018,SC2018}
%SC2018,
discuss the extension and adaptation of ISO-26262, and \cite{burton_making_2017}
%concerns 
considers
functional insufficiencies in the perception functions based on DL. 
%While 
Additionally,
\cite{picardi_pattern_2019,picardi_perspectives_2019} explores safety case patterns that are reusable for DL in the context of medical applications. While, in \cite{osborne_uas_2019}, safety case approach is reviewed as useful for assuring the safety of drones.

Formal verification  \cite{HKWW2017,katz2017reluplex,xiang2017output,GMDTCV2018,LM2017,wicker2018feature,RHK2018,wu2018game,ruan2018global,LLYCH2018} and 
coverage-guided testing \cite{sun2018concolic,PCYJ2017,sun2018testing-b,ma2018deepgauge,SHKSHA2019,sun2018concolicb} currently form the two major classes of V\&V techniques for DL, from which a collection of evidence may be obtained that supports the partial prior knowledge. The readers are referred to a recent survey \cite{Huangsurvey2018} for the introduction and summarisation of the techniques. 

\section{Discussions, Conclusions and Future  Work}\label{sec-conclusions}

In this paper, we present a novel safety argument framework for DNNs 
%via 
using
probabilistic risk assessment, mainly 
%concern 
considering
quantitative reliability claims, generalising 
%the 
this
idea to other safety related properties. We emphasise the use of probabilistic measures to describe the inherent uncertainties of DNNs in safety arguments, and conduct Bayesian inference to strengthen the top-level claims from safe operational data 
%by continuously 
through to continuous
monitoring after deployment.

Bayesian inference requires prior knowledge, so we propose a novel view by (i) decomposing the DNN generalisation error into 
%a few 
a composition of distinct
errors and (ii) 
%trying 
try to map each lifecycle activity to the reduction of 
%some 
these
errors. Although we have shown an example of obtaining priors from robustness verification of DNNs, it is %still hard 
non-trivial
(and identified as an open challenge) to establish a quantitative link between other lifecycle activities to the generalisation error. Expert judgement and past experience (e.g., a repository on DNNs developed by similar lifecycle activities) seem to be inevitable in overcoming such difficulties. %which is not surprising due to the lessons learnt from other similar approaches.

Thanks to the CBI approach -- Bayesian inference with limited and partial prior knowledge -- even with sparse prior information (e.g., a single confidence bound on the generalisation error obtained from robustness verification), we can still 
%do 
apply
probabilistic inference given the operational data. Whenever there are sound arguments to obtain additional partial prior knowledge, CBI can incorporate them as well, and reduce the conservatism in the reasoning \cite{bishop_toward_2011}. On the other hand, CBI as a type of proven-in-use/field-testing argument has some of the fundamental limitations as highlighted in \cite{KKB2019,johnson_increasing_2018}, for which we have identified on-going research towards potential solutions. 

We concur with \cite{KKB2019} that, despite the dangerous pitfalls for various existing safety arguments, credible safety cases require a heterogeneous approach. Our new quantitative safety case framework provides a novel supplementary approach to existing frameworks rather than replace them. We plan to conduct  concrete case studies %to exam the soundness of our work 
and continue to work on the open challenges identified.

%{\footnotesize\setlength{\parskip}{1em}\textbf{Acknowledgement}
\subsubsection{Acknowledgements \& disclaimer.}
This work is supported by the UK EPSRC (through the Offshore Robotics for Certification of Assets [EP/R026173/1] and its PRF project COVE, and End-to-End Conceptual Guarding of Neural Architectures [EP/T026995/1]) and the
UK Dstl (through projects on Test Coverage Metrics for Artificial Intelligence). Xingyu Zhao and Alec Banks’ contribution to the work is partially supported through Fellowships at the Assuring Autonomy International Programme.
%}

This document is an overview of UK MOD (part) sponsored research and is released for informational purposes only. The contents of this document should not be interpreted as representing the views of the UK MOD, nor should it be assumed that they reflect any current or future UK MOD policy. The information contained in this document cannot supersede any statutory or contractual requirements or liabilities and is offered without prejudice or commitment.
%\newline 
%\indent
Content includes material subject to \textcopyright~Crown copyright (2018), Dstl. This material is licensed under the terms of the Open Government Licence except where otherwise stated. To view this licence, visit \url{http://www.nationalarchives.gov.uk/doc/open-government-licence/version/3} or write to the Information Policy Team, The National Archives, Kew, London TW9 4DU, or email: psi@nationalarchives.gsi.gov.uk.
%\par}

%\begin{itemize}
%	\item as an extension/supplementary arguments to qualitative safety arguments
%	\item  Soundness needs concrete case studies.
%	\item incomplete list of properties and open challenges.
%	\item other properties do not necessarily have to use CBI for statistical inference.
%\end{itemize}

%\begin{theorem}
%This is a sample theorem. The run-in heading is set in bold, while
%the following text appears in italics. Definitions, lemmas,
%propositions, and corollaries are styled the same way.
%\end{theorem}
%
% the environments 'definition', 'lemma', 'proposition', 'corollary',
% 'remark', and 'example' are defined in the LLNCS documentclass as well.
%

 \bibliographystyle{splncs04}
 \bibliography{references}
%Author, F.: Article title. Journal \textbf{2}(5), 99--110 (2016)
%Author, F., Author, S.: Title of a proceedings paper. In: Editor,
%F., Editor, S. (eds.) CONFERENCE 2016, LNCS, vol. 9999, pp. 1--13.
%Springer, Heidelberg (2016). \doi{10.10007/1234567890}
%
%\end{thebibliography}
\end{document}

%% file: macro.tex
\newcommand{\commentout}[1]{}

\newcommand{\distance}[2]{ ||#1||_{#2}}

\newcommand{\network}{{\cal N}}

\newcommand{\networks}{{\tt N}}
\newcommand{\inputdomain}{{\tt X}}
\newcommand{\outputdomain}{{\tt Y}}

\newcommand{\dist}{{\cal D}}

\newcommand{\explain}{{\tt expl}}

\newcommand{\varX}{{\cal X}}
\newcommand{\varY}{{\cal Y}}
\newcommand{\varZ}{{\cal Z}}

%% file: main.bbl
\begin{thebibliography}{10}
\providecommand{\url}[1]{\texttt{#1}}
\providecommand{\urlprefix}{URL }
\providecommand{\doi}[1]{https://doi.org/#1}

\bibitem{Abadi_2016}
Abadi, M., Chu, A., Goodfellow, I., McMahan, H.B., Mironov, I., Talwar, K.,
  Zhang, L.: Deep learning with differential privacy. ACM SIGSAC CCS’16
  (2016)

\bibitem{alves_considerations_2018}
Alves, E., Bhatt, D., Hall, B., Driscoll, K., Murugesan, A., Rushby, J.:
  Considerations in assuring safety of increasingly autonomous systems.
  Technical {Report} NASA/CR-2018-220080, NASA (Jul 2018)

\bibitem{asaadi_towards_2019}
Asaadi, E., Denney, E., Pai, G.: Towards quantification of assurance for
  learning-enabled components. In: EDCC'19. pp. 55--62. IEEE, Naples, Italy
  (2019)

\bibitem{ashmore_assuring_2019}
Ashmore, R., Calinescu, R., Paterson, C.: Assuring the machine learning
  lifecycle: {Desiderata}, methods, and challenges. arXiv preprint
  arXiv:1905.04223  (2019)

\bibitem{bagnall_certifying_2019}
Bagnall, A., Stewart, G.: Certifying the true error: {Machine} learning in
  {Coq} with verified generalization guarantees. In: AAAI2019. vol.~33, pp.
  2662--2669 (2019)

\bibitem{barocas-hardt-narayanan}
Barocas, S., Hardt, M., Narayanan, A.: Fairness and Machine Learning.
  fairmlbook.org (2019), \url{http://www.fairmlbook.org}

\bibitem{bishop_conservative_2014}
Bishop, P., Bloomfield, R., Littlewood, B., Popov, P., Povyakalo, A., Strigini,
  L.: A conservative bound for the probability of failure of a 1-out-of-2
  protection system with one hardware-only and one software-based protection
  train. Reliability Engineering \& System Safety  \textbf{130},  61--68 (2014)

\bibitem{bishop_toward_2011}
Bishop, P., Bloomfield, R., Littlewood, B., Povyakalo, A., Wright, D.: Toward a
  formalism for conservative claims about the dependability of software-based
  systems. IEEE Transactions on Software Engineering  \textbf{37}(5),  708--717
  (2011)

\bibitem{bishop_deriving_2017}
Bishop, P., Povyakalo, A.: Deriving a frequentist conservative confidence bound
  for probability of failure per demand for systems with different operational
  and test profiles. Reliability Engineering \& System Safety  \textbf{158},
  246--253 (2017)

\bibitem{BKCF2019}
{Bloomfield}, R., {Khlaaf}, H., {Ryan Conmy}, P., {Fletcher}, G.: Disruptive
  innovations and disruptive assurance: Assuring machine learning and autonomy.
  Computer  \textbf{52}(9),  82--89 (2019)

\bibitem{bloomfield_confidence:_2007}
Bloomfield, R.E., Littlewood, B., Wright, D.: Confidence: {Its} role in
  dependability cases for risk assessment. In: DSN2007. pp. 338--346. IEEE,
  Edinburgh, UK (2007)

\bibitem{bloomfield_safety_2010}
Bloomfield, R., Bishop, P.: Safety and assurance cases: past, present and
  possible future -- an {Adelard} perspective. In: Dale, C., Anderson, T.
  (eds.) Making {Systems} {Safer}. pp. 51--67. Springer London, London (2010)

\bibitem{burton_making_2017}
Burton, S., Gauerhof, L., Heinzemann, C.: Making the case for safety of machine
  learning in highly automated driving. In: SafeComp'17. {LNCS}, vol. 10489,
  pp. 5--16. Springer, Cham (2017)

\bibitem{burton_confidence_2019}
Burton, S., Gauerhof, L., Sethy, B.B., Habli, I., Hawkins, R.: Confidence
  arguments for evidence of performance in machine learning for highly
  automated driving {Functions}. In: SafeComp'19. {LNCS}, vol. 11699, pp.
  365--377. Springer, Cham (2019)

\bibitem{chen_diversity_2016}
Chen, L., May, J.H.R.: A diversity model based on failure distribution and its
  application in safety cases. IEEE Tran. on Reliability  \textbf{65}(3),
  1149--1162 (2016)

\bibitem{denney_towards_2011}
Denney, E., Pai, G., Habli, I.: Towards measurement of confidence in safety
  cases. In: {Int.} {Symp.} on {Empirical} {Software} {Engin.} and
  {Measurement}. pp. 380--383 (2011)

\bibitem{du_gradient_2018}
Du, S.S., Lee, J.D., Li, H., Wang, L., Zhai, X.: Gradient descent finds global
  minima of deep neural networks. arXiv e-prints p. arXiv:1811.03804 (Nov 2018)

\bibitem{ferrando_verifying_2018}
Ferrando, A., Dennis, L.A., Ancona, D., Fisher, M., Mascardi, V.: Verifying and
  validating autonomous systems: {Towards} an integrated approach. In: Runtime
  {Verification}. pp. 263--281. Springer, Cham (2018)

\bibitem{fukunaga_introduction_2013}
Fukunaga, K.: Introduction to statistical pattern recognition. Elsevier (2013)

\bibitem{galves_rare_1998}
Galves, A., Gaudel, M.: Rare events in stochastic dynamical systems and
  failures in ultra-reliable reactive programs. In: FTCS'98. pp. 324--333.
  Munich, DE (1998)

\bibitem{GMDTCV2018}
{Gehr}, T., {Mirman}, M., {Drachsler-Cohen}, D., {Tsankov}, P., {Chaudhuri},
  S., {Vechev}, M.: Ai2: Safety and robustness certification of neural networks
  with abstract interpretation. In: IEEE Symposium on Security and Privacy
  (SP). pp. 3--18 (2018)

\bibitem{he_control_2019}
He, F., Liu, T., Tao, D.: Control batch size and learning rate to generalize
  well: {Theoretical} and empirical evidence. In: NIPS'19, pp. 1141--1150
  (2019)

\bibitem{Huangsurvey2018}
Huang, X., Kroening, D., Ruan, W., Sharp, J., Sun, Y., Thamo, E., Wu, M., Yi,
  X.: A survey of safety and trustworthiness of deep neural networks. arXiv
  preprint arXiv:1812.08342  (2018)

\bibitem{HKWW2017}
Huang, X., Kwiatkowska, M., Wang, S., Wu, M.: Safety verification of deep
  neural networks. In: CAV'17. {LNCS}, vol. 10426, pp. 3--29. Springer, Cham
  (2017)

\bibitem{ishikawa_continuous_2018}
Ishikawa, F., Matsuno, Y.: Continuous argument engineering: {Tackling}
  uncertainty in machine learning based systems. In: Gallina, B., Skavhaug, A.,
  Schoitsch, E., Bitsch, F. (eds.) SafeComp'18. {LNCS}, vol. 11094, pp. 14--21.
  Springer, Cham (2018)

\bibitem{johnson_increasing_2018}
{Johnson, C. W.}: The increasing risks of risk assessment: {On} the rise of
  artificial intelligence and non-determinism in safety-critical systems. In:
  the 26th {Safety}-{Critical} {Systems} {Symposium}. p.~15. Safety-Critical
  Systems Club, York, UK. (2018)

\bibitem{katz2017reluplex}
Katz, G., Barrett, C., Dill, D.L., Julian, K., Kochenderfer, M.J.: Reluplex:
  {An} efficient {SMT} solver for verifying deep neural networks. In: CAV'17.
  {LNCS}, vol. 10426, pp. 97--117. Springer, Cham (2017)

\bibitem{kelly_arguing_1999}
Kelly, T.P.: Arguing safety: {A} systematic approach to managing safety cases.
  {PhD} {Thesis}, University of York (1999)

\bibitem{KKB2019}
Koopman, P., Kane, A., Black, J.: Credible autonomy safety argumentation. In:
  27th {Safety}-{Critical} {Sys.} {Symp.} Safety-Critical Systems Club,
  Bristol, UK (2019)

\bibitem{LLYCH2018}
Li, J., Liu, J., Yang, P., Chen, L., Huang, X., Zhang, L.: Analyzing deep
  neural networks with symbolic propagation: {Towards} higher precision and
  faster verification. In: Chang, B.Y.E. (ed.) Static {Analysis}. {LNCS}, vol.
  11822, pp. 296--319. Springer International Publishing, Cham (2019)

\bibitem{littlewood_reasoning_2012}
Littlewood, B., Rushby, J.: Reasoning about the reliability of diverse
  two-channel systems in which one channel is ``possibly perfect''. TSE
  \textbf{38}(5),  1178--1194 (2012)

\bibitem{littlewood_validation_2011}
Littlewood, B., Strigini, L.: '{Validation} of ultra-high dependability...' -
  20 years on. Safety Systems, Newsletter of the Safety-Critical Systems Club
  \textbf{20}(3) (2011)

\bibitem{littlewood_conceptual_1989}
Littlewood, B., Miller, D.R.: Conceptual modeling of coincident failures in
  multiversion software. IEEE Tran. on Software Engineering  \textbf{15}(12),
  1596--1614 (1989)

\bibitem{littlewood_conservative_2013}
Littlewood, B., Povyakalo, A.: Conservative bounds for the pfd of a 1-out-of-2
  software-based system based on an assessor's subjective probability of ``not
  worse than independence''. IEEE Tran. on Soft. Eng.  \textbf{39}(12),
  1641--1653 (2013)

\bibitem{littlewood_reliability_2020}
Littlewood, B., Salako, K., Strigini, L., Zhao, X.: On reliability assessment
  when a software-based system is replaced by a thought-to-be-better one.
  Reliability Engineering \& System Safety  \textbf{197},  106752 (2020)

\bibitem{littlewood_use_2007}
Littlewood, B., Wright, D.: The use of multilegged arguments to increase
  confidence in safety claims for software-based systems: a study based on a
  {BBN} analysis of an idealized example. IEEE Transactions on Software
  Engineering  \textbf{33}(5) (2007)

\bibitem{LM2017}
Lomuscio, A., Maganti, L.: An approach to reachability analysis for
  feed-forward {ReLU} neural networks. arXiv preprint arXiv:1706.07351  (2017)

\bibitem{ma2018deepgauge}
Ma, L., Juefei{-}Xu, F., Sun, J., Chen, C., Su, T., Zhang, F., Xue, M., Li, B.,
  Li, L., Liu, Y., Zhao, J., Wang, Y.: {DeepGauge}: Comprehensive and
  multi-granularity testing criteria for gauging the robustness of deep
  learning systems. In: ASE2018. pp. 120--131. Montpellier, France (2018)

\bibitem{matsuno_tackling_2019}
Matsuno, Y., Ishikawa, F., Tokumoto, S.: Tackling uncertainty in safety
  assurance for machine learning: {Continuous} argument engineering with
  attributed tests. In: SafeComp'19. {LNCS}, vol. 11699, pp. 398--404.
  Springer, Cham (2019)

\bibitem{Micouin2008}
Micouin, P.: Toward a property based requirements theory: System requirements
  structured as a semilattice. Systems Engineering  \textbf{11}(3),  235--245
  (2008)

\bibitem{musa_operational_1993}
Musa, J.D.: Operational profiles in software-reliability engineering. IEEE
  Software  \textbf{10}(2),  14--32 (Mar 1993)

\bibitem{ohagan_uncertain_2006}
O'Hagan, A., Buck, C.E., Daneshkhah, A., Eiser, J.R., Garthwaite, P.H.,
  Jenkinson, D.J., Oakley, J.E., Rakow, T.: Uncertain judgements: {Eliciting}
  experts' probabilities. John Wiley \& Sons (2006)

\bibitem{osborne_uas_2019}
Osborne, M., Lantair, J., Shafiq, Z., Zhao, X., Robu, V., Flynn, D., Perry, J.:
  {UAS} operators safety and reliability survey: {Emerging} technologies
  towards the certification of autonomous {UAS}. In: ICSRS'19. pp. 203--212.
  IEEE, Rome (2019)

\bibitem{PCYJ2017}
Pei, K., Cao, Y., Yang, J., Jana, S.: {DeepXplore}: Automated whitebox testing
  of deep learning systems. In: SOSP'17. pp. 1--18. ACM, New York, NY, USA
  (2017)

\bibitem{picardi_perspectives_2019}
Picardi, C., Habli, I.: Perspectives on assurance case development for retinal
  disease diagnosis using deep learning. In: Riaño, D., Wilk, S., ten Teije,
  A. (eds.) AIME'19. {LNCS}, vol. 11526, pp. 365--370. Springer, Cham (2019)

\bibitem{picardi_pattern_2019}
Picardi, C., Hawkins, R., Paterson, C., Habli, I.: A pattern for arguing the
  assurance of machine learning in medical diagnosis systems. In: SafeComp'19.
  {LNCS}, vol. 11698, pp. 165--179. Springer, Cham (2019)

\bibitem{Ponti2011}
{Ponti Jr.}, M.P.: Combining classifiers: From the creation of ensembles to the
  decision fusion. In: SIBGRAPI'11. pp. 1--10. IEEE, Alagoas, Brazil (2011)

\bibitem{RHK2018}
Ruan, W., Huang, X., Kwiatkowska, M.: Reachability analysis of deep neural
  networks with provable guarantees. In: IJCAI2018. pp. 2651--2659 (2018)

\bibitem{ruan2018global}
Ruan, W., Wu, M., Sun, Y., Huang, X., Kroening, D., Kwiatkowska, M.: Global
  robustness evaluation of deep neural networks with provable guarantees for
  the hamming distance. In: IJCAI2019. pp. 5944--5952 (2019)

\bibitem{rudolph_consistent_2018}
Rudolph, A., Voget, S., Mottok, J.: A consistent safety case argumentation for
  artificial intelligence in safety related automotive systems. In: {ERTS}'18
  (2018)

\bibitem{rushby_software_2009}
Rushby, J.: Software verification and system assurance. In: 7th {Int.} {Conf.}
  on {Software} {Engineering} and {Formal} {Methods}. pp. 3--10. IEEE, Hanoi,
  Vietnam (2009)

\bibitem{s_toulmin_uses_1958}
{S. Toulmin}: The {Uses} of {Argument}. Cambridge University Press (1958)

\bibitem{SC2018}
Salay, R., Czarnecki, K.: Using machine learning safely in automotive software:
  An assessment and adaption of software process requirements in {ISO} 26262.
  arXiv preprint arXiv:1808.01614  (2018)

\bibitem{SS2020b}
Schwalbe, G., Schels, M.: Concept enforcement and modularization as methods for
  the {ISO} 26262 safety argumentation of neural networks. In: ERTS'20 (2020)

\bibitem{SS2020}
Schwalbe, G., Schels, M.: A survey on methods for the safety assurance of
  machine learning based systems. In: ERTS'20 (2020)

\bibitem{Sha2001}
Sha, L.: Using simplicity to control complexity. IEEE Software  \textbf{18}(4),
   20--28 (2001)

\bibitem{strigini_testing_1996}
Strigini, L.: On testing process control software for reliability assessment:
  the effects of correlation between successive failures. Software Testing,
  Verification and Reliability  \textbf{6}(1),  33--48 (1996)

\bibitem{strigini_software_2013}
Strigini, L., Povyakalo, A.: Software fault-freeness and reliability
  predictions. In: SafeComp'13. {LNCS}, vol.~8153, pp. 106--117. Springer,
  Berlin, Heidelberg (2013)

\bibitem{sun2018testing-b}
Sun, Y., Huang, X., Kroening, D., Sharp, J., Hill, M., Ashmore, R.: Structural
  test coverage criteria for deep neural networks. ACM Transactions on Embedded
  Computing Systems  \textbf{18}(5s) (2019)

\bibitem{SHKSHA2019}
Sun, Y., Huang, X., Kroening, D., Sharp, J., Hill, M., Ashmore, R.: Structural
  test coverage criteria for deep neural networks. In: ICSE'19--Companion. pp.
  320--321. IEEE Press, Piscataway, NJ, USA (2019)

\bibitem{sun2018concolic}
Sun, Y., Wu, M., Ruan, W., Huang, X., Kwiatkowska, M., Kroening, D.: Concolic
  testing for deep neural networks. In: ASE'18. pp. 109--119. ACM, New York,
  NY, USA (2018)

\bibitem{sun2018concolicb}
Sun, Y., Wu, M., Ruan, W., Huang, X., Kwiatkowska, M., Kroening, D.:
  Deepconcolic: Testing and debugging deep neural networks. In:
  ICSE'19--companion. pp. 111--114. Montreal, QC, Canada (2019)

\bibitem{wang_confidence_2017}
Wang, R., Guiochet, J., Motet, G.: Confidence assessment framework for safety
  arguments. In: Tonetta, S., Schoitsch, E., Bitsch, F. (eds.) SafeComp'17.
  {LNCS}, vol. 10488, pp. 55--68. Springer International Publishing, Cham
  (2017)

\bibitem{wicker2018feature}
Wicker, M., Huang, X., Kwiatkowska, M.: Feature-guided black-box safety testing
  of deep neural networks. In: Beyer, D., Huisman, M. (eds.) Tools and
  {Algorithms} for the {Construction} and {Analysis} of {Systems}. {LNCS}, vol.
  10805, pp. 408--426. Springer International Publishing, Cham (2018)

\bibitem{wu2018game}
Wu, M., Wicker, M., Ruan, W., Huang, X., Kwiatkowska, M.: A game-based
  approximate verification of deep neural networks with provable guarantees.
  Theoretical Computer Science  \textbf{807},  298 -- 329 (2020)

\bibitem{xiang2017output}
Xiang, W., Tran, H.D., Johnson, T.T.: Output reachable set estimation and
  verification for multi-layer neural networks. IEEE Transactions on Neural
  Networks and Learning Systems  \textbf{29},  5777--5783 (2018)

\bibitem{zhao_modeling_2017}
Zhao, X., Littlewood, B., Povyakalo, A., Strigini, L., Wright, D.: Modeling the
  probability of failure on demand (pfd) of a 1-out-of-2 system in which one
  channel is ``quasi-perfect''. Reliability Engineering \& System Safety
  \textbf{158},  230--245 (2017)

\bibitem{zhao_assessing_2019}
Zhao, X., Robu, V., Flynn, D., Salako, K., Strigini, L.: Assessing the safety
  and reliability of autonomous vehicles from road testing. In: the 30th {Int}.
  {Symp}. on {Software} {Reliability} {Engineering}. pp. 13--23. IEEE, Berlin,
  Germany (2019)

\bibitem{zhao_new_2012}
Zhao, X., Zhang, D., Lu, M., Zeng, F.: A new approach to assessment of
  confidence in assurance cases. In: Ortmeier, F., Daniel, P. (eds.) Computer
  {Safety}, {Reliability}, and {Security}. {LNCS}, vol.~7613, pp. 79--91.
  Springer, Berlin, Heidelberg (2012)

\end{thebibliography}
